# ASRL:A robust loss function with potential for development

1st CHENYU HUI*(Student Member,IEEE)
Xi'an Jiaotong University
Faculty of Microelectronics Science and Engineering
Xi'an Shannxi 710049 China
*(corresponding author)2687123206@stu.xjtu.edu.cn

2nd ANRAN ZHANG(Student Member,IEEE)
Xi'an Jiaotong University
Faculty of Electronic and Information Engineering
Xi'an Shannxi 710049 china
179404529@stu.xjtu.edu.cn

3rd XINTONG LI
Xi'an Jiaotong University
Faculty of Microelectronics Science and Engineering
Xi'an Shannxi 710049 china
2223212033@stu.xjtu.edu.cn

*Abstract*—In this article, we proposed a partition-wise robust loss function (ASRL -Adapative segmented robust loss )based on the previous robust loss function. The characteristics of this loss function are that it achieves high robustness and a wide range of applicability through partition-wise design and adaptive parameter adjustment. Finally, the advantages and development potential of this loss function were verified by applying this loss function to the XGBoost and using five different datasets (with different dimensions, different sample numbers, and different fields) to compare with the XGBoost using other loss functions.

The results of multiple experiments have proven the advantages of ASRL in MSE, MAE, $R^2$, etc. ASRL's dynamic segmentation design and adaptive threshold make it more robust and can be applied to more fields, such as as a loss function for multimodal learning and reinforcement learning, and has a large room for development.The implementation code repository github link in this paper is:ASRLCODE

## I. INTRODUCTION

In regression prediction of machine learning, the loss function is the core tool to measure the difference between the model prediction value and the true value. Its role runs through the entire process of model training, optimization and evaluation.

Classic and commonly used loss functions include MSE (MSE is one of the most commonly used loss functions in regression tasks, also known as mean square error, which is simple and fast), MAE (also known as mean absolute error, which is robust to outliers and has stable gradients, but may converge slowly), and Huber Loss (which combines the advantages of L1 and L2 losses, which are MAE and MSE respectively, and are robust to outliers. [1]).Smooth L1 Loss (It proposes Smooth L1 loss in target detection to solve the problem that L2 loss is sensitive to outliers. [2])

Representative loss functions proposed in recent years include **Adapative Robust Loss** (a generalized robust loss function is proposed, which dynamically adjusts the sensitivity of loss to outliers through parameters [3]) and **Charbonnier Loss** (which improves L1 loss into a smooth form and is mostly used in tasks such as image restoration [4]); **Uncertainty-weighted Loss** (which focuses on multi-task and multi-target losses [6]) and **Grad Norm** (which dynamically adjusts multi-task loss weights through gradient normalization to solve the problem of unbalanced convergence speed between tasks [7]) that focus on improving the robustness of loss functions.

In the paper "A General and Adaptive Robust Loss Function" , a single loss function is proposed A continuous-valued parameter in this general adaptive robust loss function can be set, which enables it to generalize algorithms built around fixed robust losses with a new "robustness" hyperparameter that can be adjusted or annealed to improve performance. In this paper, the hyperparameters are α and c (α is a shape parameter used to control the robustness of the function, and c is a scale parameter used to control the performance of the loss function near x=0). The main focus of this paper is the application of loss functions in image algorithms. [3]

This article refers to the design of the adaptive loss function in "A General and Adaptive Robust Loss Function", applies the adaptive loss function to the field of regression prediction, and associates the hyperparameters (quantile low and quantile high) with the characteristics of the sample during the design process. By combining the loss function with the probability distribution , the characteristics of the probability distribution are extracted and used as the boundary basis for the segmented processing of the loss function. This probabilistic framework not only avoids the instability caused by direct optimization of α (such as α approaching negative infinity to completely ignore anomalies), but also provides theoretical guarantees.To clearly demonstrate the specific and robust nature of ASRL as an adaptive piecewise function, we create the following diagram, which contains the ASRL loss function for three residual segments.The three residual intervals in the diagram below each indicate the corresponding loss function.(Fig1)

In general, the innovations of ASRL are:

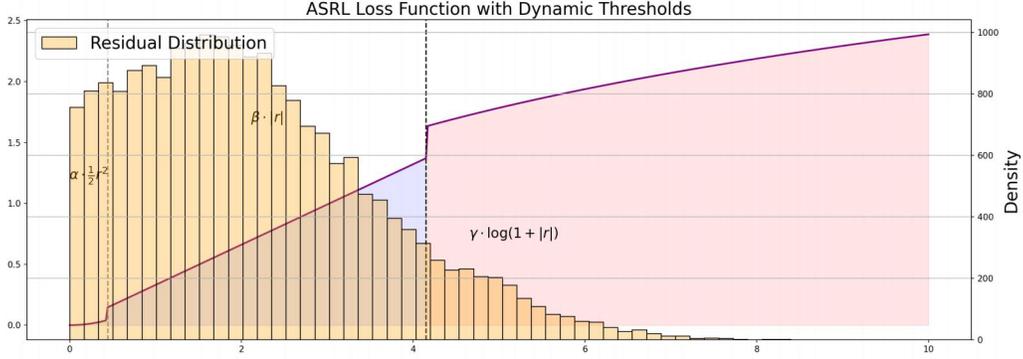

Fig. 1. ASRL(Adapative segmented robust loss )Loss function visualization

- Using segmented dynamic threshold, ASRL divides the residuals into three areas: small, medium, and large by dynamically calculating the quantile threshold of the sample data residuals.
- Using segmented loss weight, different loss weights are used in different areas. For example, logarithmic loss is used in large residual areas. The gradient decays as the residual increases, suppressing the influence of outliers.
- Correlation with probability distributions: Correlation of the loss function with generalized probability distributions (such as piecewise Gaussian mixture distributions) automatically optimizes parameters by maximizing the likelihood.

In the following part of this article, we first explain the mathematical principles of ASRL and the rationality of its design. Then in the experimental part, we demonstrate the advantages and development potential of the ASRL loss function by testing five datasets with different numbers of features, different fields and different numbers of samples and performing visualization operations.

## II. PROBLEM DEFINITION

In the task of machine learning (whose goal is to predict one or more continuous numerical variables), the loss function is the core driving force of model optimization [5], which directly determineshow the model learns the mapping relationship between features and targets from the data. In the process of model training, the loss function guides the direction of model iteration and optimization. Common loss functions in regression tasks are: MSE, MAE, Huber Loss. Based on the idea of designing the piecewise loss function of Huber Loss, this article mainly draws on the concept of hyperparameters in "A General and Adaptive Robust Loss Function" [3]and uses the relationship between the design of the partition threshold and the quantile statistical probability distribution to adaptively adjust the threshold and the weights of different loss intervals.

## III. METHODOLOGY

### A. Design of piecewise loss function

ASRL (Adaptive Segmented Robust Loss)proposed in this article is a loss function that is dynamically adjusted based on residual distribution. It balances the convergence speed and robustness of the model through segmented design. Its core idea is to divide the prediction error (residual) into different intervals, apply different loss functions to each interval, and dynamically adjust the parameters through an adaptive mechanism. [9]

First, regarding the design of the partition loss function, in the small residual area ($|\text{Residual}| \leq \delta_1$), square loss is used to ensure fast convergence.The gradient formula for squared loss is as follows [15]:

$$\frac{\partial L}{\partial F} = -\alpha(y - F) \tag{1}$$

In the above formula, y refers to the actual value, F refers to the current predicted value, and the value of$|y - F|$ is the residual. The size of its gradient is proportional to the residual y-F. The smaller the residual, the smaller the gradient.The size of its gradient is proportional to the residual y- F. The smaller the residual, the smaller the gradient.When the residual is large, the gradient is large, the parameters are updated faster, and the optimal point is approached quickly.When the residual is small,the gradient automatically decays to avoid parameter oscillation and stabilize convergence.

In the medium residual region($\delta_1 < |\text{Residual}| \leq \delta_2$), the absolute loss is used to suppress medium noise.The gradient formula for the medium residual region is as follows:

$$\frac{\partial L}{\partial F} = -\beta \cdot sign(y - F) \tag{2}$$

The direction of the medium residual region 's gradient is determined by the sign of the residual (sign(y-F))(sign is a sign function used to determine the positive or negative value. The returned value is 1, -1, 0.), and its size is constant at β, regardless of the absolute value of the residual [13].

In the large residual area($|\text{Residual}| > \delta_2$), ASRL use logarithmic loss to reduce the impact of outliers.The gradient formula in the large residual region is as follows:

$$\frac{\partial L}{\partial F} = -\gamma \cdot \frac{sign(y-F)}{1+|y-F|} \quad (3)$$

In the large residual area, the gradient decays as the residual increases, suppressing the influence of outliers.

### B. Design of dynamic threshold

In ASRL, dynamic quantile threshold is used to divide the residual into three regions, namely small residual region, medium residual region and large residual region. In ASRL training, the quantile threshold is dynamically calculated according to the distribution of the current data residual, rather than a fixed setting, so that the model can adapt to different residual distributions [8]. The dynamic setting of the quantile threshold is implemented in the following pseudo code(Algorithm 1), where the input variable r is the absolute value sequence of the residual of the current batch (Batch) or training cycle (Epoch) and the preset low quantile and high quantile [12].

**Input:** Residual array **r** = [$r_1, r_2, ..., r_n$], low quantile $q_{low}$, high quantile $q_{high}$
**Output:** Quantile thresholds $\delta_1, \delta_2$

**Step 1: Compute absolute residuals**
**abs_r** ← [ |$r_i$| ∀$r_i$ ∈ **r** ] ;     // Generate absolute value array

**Step 2: Sort absolute residuals**
**sorted_r** ← sort(**abs_r**) ;     // Sort in ascending order to get
$r_{(1)} \le r_{(2)} \le ... \le r_{(n)}$

**Step 3: Compute quantile positions**
**for** q ∈ {$q_{low}$, $q_{high}$} **do**
  k ← q · (n – 1) ;     // Quantile position index (0-based)
  idx_floor ← ⌊k⌋ ;     // Integer part
  frac ← k – idx_floor ;     // Fractional part
  **if** idx_floor + 1 < n **then**
    δ ← (1–frac)·$r_{(idx\_floor+1)}$ +frac·$r_{(idx\_floor+2)}$
    ;     // Linear interpolation
  **else**
    δ ← $r_{(n)}$ ;     // Boundary case: take the maximum value
  **end**

  **if** q = $q_{low}$ **then**
    $\delta_1$ ← δ
  **else**
    $\delta_2$ ← δ
  **end**
**end**
**return** $\delta_1, \delta_2$

**Algorithm 1:** Dynamic Quantile Threshold Calculation

### C. Construction of ASRL loss function

After designing the piecewise loss function and planning the dynamic threshold, this paper presents the complete model of the ASRL loss function and the definition of its parameters.The equation of ASRL is as follows:

$$L(y, F) = \begin{cases} \alpha \cdot \frac{(y-F)^2}{2} & \text{if } |y-F| \le \delta_1 \\ \beta \cdot |y-F| & \text{if } \delta_1 < |y-F| \le \delta_2 \\ \gamma \cdot \log(1+|y-F|) & \text{if } |y-F| > \delta_2 \end{cases} \quad (4)$$

The three parameters α,β, and γ in the above formula are the weights of the small residual area, the medium residual area, and the large residual area, respectively.

$$\alpha = \frac{1}{\sigma^2 + \epsilon} \quad (5)$$

$$\beta = \frac{1}{\text{IQR}(|y - F|) + \epsilon} \quad (6)$$

$$\gamma = \frac{1}{\text{MAD}(y - F) + \epsilon} \quad (7)$$

In the calculation formula of α, $\sigma^2$ is the variance of the residual, and ϵ is a minimum value (such as $10^{-6}$) to prevent the denominator from being zero.The design principle of α is to dynamically amplify the gradient of the low-noise area through the variance, and use the quadratic convergence of the square loss to speed up the convergence.

In the calculation formula of parameter β, IQR is the difference between the third quartile (Q3) and the first quartile(Q1).The smaller the IQR, the more concentrated the data in the middle residual area is. [10]

In the calculation formula of γ, MAD is the median of the absolute deviations of the data points from the median of the data, that is:

$$\text{MAD} = \text{median}(|x_i - \text{meadian}(x)|) \quad (8)$$

In ASRL, MAD is used to calculate the weight parameter γ in the large residual region.The smaller the MAD, the more concentrated the data in the large residual area is (the fewer outliers), and the larger the γ weight is, which suppresses the gradient contribution of outliers [11].

## IV. EXPERIMENT

### A. Overview of experimental dataset

In order to test the performance of the ASRL loss function we constructed, we measured it on five different small and medium-sized datasets. These five datasets are the **California House Price Dataset** (the data comes from the 1990 California House data, which is used to predict the median house price in the region.), the **Gas CO and NOx emissions** (the data comes from a gas turbine located in Turkey for the purpose of studying flue gas emissions), **Combined Cycle Power Plant** (the dataset is collected from a power station working at full load), **Concrete Compressive Strength** (concrete compressive strength dataset), **Airfoil Self-Noise** (a dataset related to NASA's space mission).

Among the five evaluation indicators tested, MSE and MAE are mean square error and mean absolute error respectively. The smaller the value, the better. $R^2$ is the coefficient of determination, which measures the improvement of the model relative to the mean prediction. The closer its value is to 1, the better. Recall is originally used to measure the directionality of classification problems. Here it is used to measure the model's coverage of key samples or safety boundaries. Time can be used to measure the running cost of models using different loss functions.

*B. Experimental results analysis and indicator quantification*

In the test of loss function performance, we tried to make the test conditions of different loss functions the same as much as possible by using the same regression model (XGBoost), setting the same learning rate and number of iterations (100 cycles and learning rate of 0.1).

Because MSE and MAE can be used as both loss functions and evaluation indicators for regression prediction, in order to avoid confusion, in the following five tables, MSE is marked with the abbreviation (LS, which stands for Loss function) when used as the loss function of the regression model. After multiple tests, and printing out the five indicators of each test, we finally sorted out the corresponding test indicator results according to each dataset.

In the California housing price dataset, there are eight numerical features and one target: the median housing price (in US$100,000). It has more than 20,000 data samples. After analysis, it is found that the average value of the test set y is 2.05. After the experiment, the results evaluation of the four loss functions are compared as shown below(TABLE I) [14]:

TABLE I
COMPARSION BASED ON CALIFORNIA HOUSING DATASET

| Metric | Method | | | |
|---|---|---|---|---|
| | ASRL | MSE (LS) | MAE (LS) | Huber |
| MSE | 0.26 | 0.29 | 0.34 | 0.30 |
| MAE | 0.33 | 0.37 | 0.38 | 0.36 |
| R2 | 0.80 | 0.77 | 0.74 | 0.78 |
| Recall | 0.90 | 0.90 | 0.90 | 0.89 |
| Time (s) | 9.80 | 6.00 | 6.40 | 6.16 |

In the concrete comprehensive strength datasets, there are eight eigenvalues related to concrete and one target value: the comprehensive strength of concrete. The dataset has 1030 samples. Through data analysis, it is known that the average value of y in the test set is 35.6. And the results evaluation of the XGBoost regression model using different loss functions are as follows(TABLE II):

In the Gas Turbine CO and NOx Emission DataSet used, there are 12 feature variables, and the target variable to be predicted is turbine energy yield. The number of samples is more than 36,000, average value of y in the test set is 135, and

TABLE II
COMPARSION BASED ON CONCRETE DATASET

| Metric | Method | | | |
|---|---|---|---|---|
| | ASRL | MSE (LS) | MAE (LS) | Huber |
| MSE | 18.3 | 20.9 | 22.1 | 18.8 |
| MAE | 3.2 | 3.5 | 3.5 | 2.9 |
| R2 | 0.93 | 0.92 | 0.92 | 0.93 |
| Recall | 0.84 | 0.88 | 0.88 | 0.84 |
| Time (s) | 0.46 | 0.22 | 0.26 | 0.28 |

the results evaluation of the XGBoost regression model using different loss functions are as follows(TABLE III):

TABLE III
COMPARSION BASED ON GAS DATASET

| Metric | Method | | | |
|---|---|---|---|---|
| | ASRL | MSE (LS) | MAE (LS) | Huber |
| MSE | 2.48 | 2.88 | 5.79 | 3.57 |
| MAE | 1.16 | 1.18 | 1.54 | 1.17 |
| R2 | 0.99 | 0.98 | 0.97 | 0.98 |
| Recall | 0.97 | 0.94 | 0.83 | 0.89 |
| Time (s) | 0.81 | 0.75 | 0.73 | 0.73 |

In the Combined Cycle Power Plant dataset, the number of features of the samples is 4, the target value of the samples is net hourly electrical energy output (EP), the number of samples is more than 9,500, and after data analysis, the average value of y in the test set is 454. The results evaluation of the XGBoost regression model using different loss functions are as follows(TABLE IV):

TABLE IV
COMPARSION BASED ON POWER PLANT DATASET

| Metric | Method | | | |
|---|---|---|---|---|
| | ASRL | MSE (LS) | MAE (LS) | Huber |
| MSE | 13.87 | 14.65 | 15.42 | 14.57 |
| MAE | 2.75 | 2.94 | 2.98 | 2.91 |
| R2 | 0.95 | 0.94 | 0.94 | 0.94 |
| Recall | 0.94 | 0.93 | 0.93 | 0.93 |
| Time (s) | 3.39 | 1.36 | 1.45 | 1.69 |

In the Airfoil Self-Noise dataset, the number of sample features is 5, the target value of regression prediction is scaled-sound-pressure, and the number of samples is more than 1,500. After data analysis, the average y value of the test set is 124.6. The results of XGBoost regression prediction using different loss functions are evaluated as follows(TABLE V):

*C. Experimental data visualization*

In order to visualize the prediction performance of XGBoost using the ASRL loss function, based on the California house price dataset mentioned above, some predicted values are compared with the original values. The results are as follows(Fig 2):

In the bar chart below(Fig 3), by comparing the MSE (root mean square error) indicators of the five loss functions ASRL,

TABLE V
COMPARSION BASED ON NOISE DATASET

| Metric | Method | | | |
|---|---|---|---|---|
| | ASRL | MSE (LS) | MAE (LS) | Huber |
| MSE | 3.56 | 3.79 | 4.88 | 3.99 |
| MAE | 1.42 | 1.48 | 1.69 | 1.53 |
| $R_2$ | 0.93 | 0.92 | 0.90 | 0.91 |
| Recall | 0.96 | 0.95 | 0.95 | 0.96 |
| Time (s) | 0.38 | 0.10 | 0.24 | 0.28 |

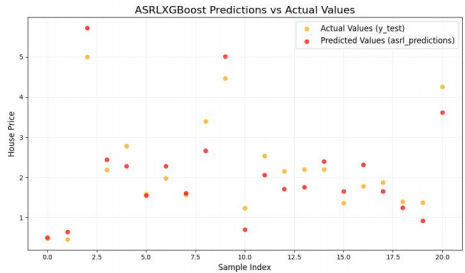

Fig. 2. Comparison of predict and original

MSE (as Loss function), MAE (as Loss function), and Huber Loss, we can clearly see the advantage of ASRL, that is, its MSE is usually lower.

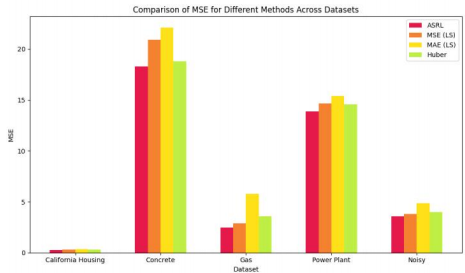

Fig. 3. Compare of four loss functions

## V. CONCLUSIONS

ASRL's advantage as a loss function is obvious, that is, it can dynamically and flexibly adjust the threshold according to different residual values and match different loss functions under different thresholds. In the actual training process, ASRL always converges faster under the same training cycle and learning rate background.

ASRL shows broad application prospects in complex data scenarios. Its core potential comes from the flexible adaptability of dynamic quantile thresholds and adaptive weight mechanisms to multimodal noise, outliers and asymmetric distributions. On the technical level, ASRL dynamically divides residual regions (small, medium and large errors) by quantiles, and automatically adjusts the weights of each region based on statistics such as variance and interquartile range. It can not only retain the fast convergence of small residual regions, but also suppress the interference of outliers in large residual regions. ASRL can automatically identify and weaken the gradient contribution of fraudulent samples without presetting a threshold. With the popularization of edge computing and the Internet of Things, ASRL's low parameter requirements and high robustness further adapt to resource-constrained scenarios. In addition, in emerging fields such as multimodal learning and reinforcement learning [16].

ACKNOWLEDGMENT

We should thank Hongyi Duan (dannhiroaki.github.io)for his guidance and help in this paper. As a student with rich scientific research experience, he played a great role in helping us revise and improve the paper.

In the experimental part of this article, we used multiple online open source datasets. In particular, we would like to thank the UCI machine learning repositories and the collation and publishers of the datasets.